\documentclass{article}

\usepackage{makeidx}  % allows for indexgeneration

\usepackage{graphicx}
\usepackage{cite}

\usepackage{color}
\usepackage{amsmath}
\usepackage{epsfig}
\usepackage{url}

\usepackage{bm}

\newtheorem{lem}{Lemma}

\usepackage{amssymb}
\usepackage{amsfonts}

\newcommand{\qed}{\fbox{\rule[1pt]{0pt}{0pt}}}
\newcommand{\bsquare}{\hbox{\rule{6pt}{6pt}}}

\newcommand{\bfe}{{\bf e}}
\newcommand{\bfx}{{\bf x}}

\newcommand{\bfr}{{\bf r}}

\newcommand{\R}{{\bf R}}
\newcommand{\A}{{\bf A}}
\newcommand{\B}{{\bf B}}

\newcommand{\I}{{\bf I}}

\newcommand{\X}{{\bf X}}

\title{Joint estimation of linear non-Gaussian acyclic models}
\author{Shohei Shimizu\thanks{The Institute of Scientific and Industrial Research (ISIR), Osaka University, Mihogaoka 8-1, Ibaraki, Osaka 567-0047, Japan. Email: sshimizu@ar.sanken.osaka-u.ac.jp}, 
}
\date{}

\begin{document}

\maketitle

\begin{abstract}
A linear non-Gaussian structural equation model called LiNGAM is an identifiable model for exploratory causal analysis. Previous  methods estimate a causal ordering of variables and their connection strengths based on a single dataset. However, in many application domains, data are obtained under different conditions, that is, multiple datasets are obtained rather than a single dataset. In this paper, we present a new method to jointly estimate multiple LiNGAMs under the assumption that the models share a causal ordering but may have different connection strengths and differently distributed variables. In simulations, the new method estimates the models more accurately than estimating them separately. 
\end{abstract}

%%%%%%%%%%%%%%%%%%%%%%%%%%%%%%%%%%%%
\section{Introduction}\label{sec:intro}
%%%%%%%%%%%%%%%%%%%%%%%%%%%%%%%%%%%%
In causal discovery methods for continuous variables, 
it is typical \cite{Spirtes93book} to assume i) the data generating process of observed variables $x_i$ ($i=1,\cdots,p$) is graphically represented by a directed acyclic graph, that is,  DAG; ii) the relations are linear; iii) there is no unobserved confounding variable. 
Let us denote by $b_{ij}$ the connection strength from $x_j$ to $x_i$ and by $k(i)$ a causal order of $x_i$ in the DAG so that no later variable determines or has a directed path on any earlier variable. 
Without loss of generality, each observed variable $x_i$ is assumed to have zero mean. 
Then we have 
\begin{eqnarray}
x_i &=& \sum_{k(j)<k(i)} b_{ij}x_j + e_i,\label{eq:model1}
\end{eqnarray}
where $e_i$ are external influences that are continuous variables with zero means and non-zero variances and are mutually independent. The independence assumption between $e_i$ means that there is no latent confounding variable. 
In matrix form, the model (\ref{eq:model1}) is written as
\begin{eqnarray}
\bfx &=& \B\bfx + \bfe,\label{eq:model2}
\end{eqnarray}
where the connection strength matrix or adjacency matrix $\B$ collects $b_{ij}$ and the vectors $\bfx$ and $\bfe$ collect $x_i$ and $e_i$, respectively.  
Note that the matrix $\B$ can be permuted  to be lower triangular with all zeros on the diagonal if simultaneous equal row and column permutations are made according to a causal ordering $k(i)$ \cite{Bollen89book}.
The problem of causal discovery is now to estimate the connection strength matrix $\B$ based on data $\bfx$ only. 
Each $b_{ij}$ represents the direct causal effect of $x_j$ on $x_i$ and each $a_{ij}$, the $(i, j)$-th element of the matrix $\A=(\I-\B)^{-1}$, the total causal effect of $x_j$ on $x_i$ \cite{Hoyer07IJAR}. 
In LiNGAM \cite{Shimizu06JMLR}, external influences $e_i$ are assumed to be {\it non-Gaussian}. This makes it possible to identify a causal ordering $k(i)$ without using any background knowledge on the structure \cite{Shimizu06JMLR}.  
Once a causal ordering $k(i)$  is identified, the connection strengths $b_{ij}$ can be estimated by ordinary least squares regression \cite{Shimizu06JMLR}. 
Many works have been conducted to extend LiNGAM to more general cases \cite{Hyva10JMLR, Hoyer09NIPS, Lacerda08UAI} and apply LiNGAM and its extensions on real-world problems \cite{Faes10EMBS,Fer10EE,Sogawa10ICANN}. 

Previous methods for LiNGAM \cite{Shimizu06JMLR,Shimizu11JMLR} estimate a causal ordering $k(i)$  based on a single dataset. 
However, in many applications, data are often obtained under different conditions. 
In neuroinformatics, fMRI (functional magnetic resonance imaging) signals are frequently recorded for multiple subjects \cite{Esposito05NI, Smith11NI}. 
In bioinformatics, gene expression levels are measured under various experimental conditions \cite{Shimamura10Binfo}. 
That is, multiple datasets are obtained rather than a single dataset. 
Since such data are likely to have some common structure, it would be effective in estimation to exploit the similarities \cite{Tillman09ICML}.  
It is well known \cite{Lee82PMK,Tillman09ICML} that it could cause serious bias in estimation to naively concatenate multiple datasets into a single dataset and analyze it. 
Many sophisticated methods for such situations have been proposed in other statistical models \cite{Shimamura10Binfo,Tillman09ICML,Lee10NIPS,Lee82PMK}. 
However, to our best knowledge, no such method is yet proposed in the literature of LiNGAM. 
Thus, in this paper, we propose a method to jointly estimate such LiNGAMs that share a causal ordering but may have different connection strength matrices and differently distributed external influences. 

%%%%%%%%%%%%%%%%%%%%%%%%%%%%%%%%%%%%
\section{Model definition}
%%%%%%%%%%%%%%%%%%%%%%%%%%%%%%%%%%%%
We now define our model extending LiNGAM \cite{Shimizu06JMLR}, that is, the model~(\ref{eq:model2}) with non-Gaussian external influences, to multiple group cases.  
We assume the following data generating processes:
\begin{eqnarray}
x_i^{(g)} &=& \sum_{k(j)<k(i)} b_{ij}^{(g)}x_j^{(g)} + e_i^{(g)} \hspace{5mm} (g=1,\cdots,c), \label{eq:model3}
\end{eqnarray}
where $g$ is the group index, $c$ is the number of groups, $k(i)$ is such a causal ordering of variables that they form a DAG in each group $g$, $x_i^{(g)}$, $e_i^{(g)}$ and $b_{ij}^{(g)}$ are the observed variables, external influences, and connection strengths of Group $g$, respectively. 
External influences $e_i^{(g)}$ are  continuous {\it non-Gaussian} variables with zero means and non-zero variances and are mutually independent. 
In matrix form, the model~(\ref{eq:model3}) is written as
\begin{eqnarray}
\bfx^{(g)} &=& \B^{(g)}\bfx^{(g)} + \bfe^{(g)} \hspace{5mm} (g=1,\cdots,c). \label{eq:model}
\end{eqnarray}
The connection strength matrix $\B^{(g)}$ collects $b_{ij}^{(g)}$ and the vectors $\bfx^{(g)}$ and $\bfe^{(g)}$ collect $x_i^{(g)}$ and $e_i^{(g)}$, respectively. 
The assumption that the models shares a causal ordering $k(i)$ might be rather weak since a recent method \cite{Tillman09ICML} for conventional graphical models makes a much stronger assumption that the zero/non-zero patterns of connection strength matrices, that is,  the DAG structures, are equivalent between the groups. 

%%%%%%%%%%%%%%%%%%%%%%%%%%%%%%%%%%%%
\section{A direct estimation method for multiple LiNGAMs}\label{sec:est}
%%%%%%%%%%%%%%%%%%%%%%%%%%%%%%%%%%%%
First, we briefly review an estimation method for a single group in Subsection~\ref{sec:single} and extend it to multiple group cases in Subsection~\ref{sec:proposal}. 

%%%%%%%%%%%%%%%%%%%%%%%%%%%%%%%%%%%%
\subsection{Background: A direct method for a single group}\label{sec:single}
%%%%%%%%%%%%%%%%%%%%%%%%%%%%%%%%%%%%
In \cite{Shimizu11JMLR}, a direct estimation method called DirectLiNGAM was proposed. 
DirectLiNGAM estimates causal orders one by one and eventually a causal ordering of all the variables. 
An exogenous variables is a variable with no parents, and the corresponding row of $\B$ has all zeros. 
Therefore, an exogenous variable can be at the top of such a causal ordering that makes $\B$ lower triangular with zeros on the diagonal. 
Then we remove the effect of the exogenous variable from the other variables by regressing it out. 
We iterate this procedure until all the variables are ordered. 
Now the point is how we can find an exogenous variable. 
The following lemma of \cite{Shimizu11JMLR} shows how it is possible: 
\begin{lem}\label{lemma1}
Assume that the input data $\bfx$ strictly follows LiNGAM, that is, the model~(\ref{eq:model2}) with non-Gaussian external influences. 
This means that we assume that all the model assumptions are met and the sample size is infinite. 
Denote by $r_i^{(j)}$ the residual  when $x_i$ is regressed on $x_j$: 
$r_i^{(j)} = x_i - \frac{{\rm cov}(x_i,x_j)}{{\rm var}(x_j)}x_j$ $(i \neq j)$.
Then a variable $x_j$ is exogenous if and only if $x_j$ is independent of its residuals $r_i^{(j)}$ for all $i \neq j$. \mbox{\hfill \qed}
\end{lem} 

To evaluate independence between a variable $x_j$ and its residuals $r_i^{(j)}$ ($i \neq j$), 
we first evaluate pairwise independence between the variable and each of the residuals using a kernel-based estimator of mutual information called KGV \cite{Bach02JMLR}, which we denote by  $KGV(x_i,r_i^{(j)})$, and subsequently compute the sum of the pairwise independence measures over the residuals. 
The non-negative estimator KGV asymptotically goes to zero if and only if the variables are independent \cite{Bach02JMLR}. 
Thus we obtain the following statistic to evaluate independence between a variable $x_j$ and its residuals $r_i^{(j)}$: 
\begin{eqnarray}
T_{kernel}(x_j; U) &=& \sum_{i\in U, i \neq j} KGV(x_j, r_i^{(j)}), \label{eq:Tkernel}
\end{eqnarray}
where $U$ denotes the set of the subscripts of variables $x_i$, that is,  $U$$=$\{$1$, $\cdots$, $p$\}. 

%%%%%%%%%%%%%%%%%%%%%%%%%%%%%%%%%%%%
\subsection{A direct algorithm for multiple groups}\label{sec:proposal}
%%%%%%%%%%%%%%%%%%%%%%%%%%%%%%%%%%%%
In this subsection, we present a method to jointly estimate multiple LiNGAMs constraining their causal orderings to be  identical. 
This would enable more accurate estimation of the LiNGAMs in Eq.~(\ref{eq:model}) than estimating them separately since they share a causal ordering. 

We first extend Lemma~\ref{lemma1} to multiple-group cases:
\begin{lem}\label{lemma2}
Assume that the input data $\bfx^{(g)}$ ($g=1, \cdots, c$) strictly follow the model (\ref{eq:model}). 
Denote by $r^{(g)}_{i,(j)}$ the residual  when $x_i^{(g)}$ is regressed on $x_j^{(g)}$: 
$r^{(g)}_{i,(j)} = x_i^{(g)} - \frac{{\rm cov}(x_i^{(g)},x_j^{(g)})}{{\rm var}(x_j^{(g)})}x_j^{(g)}$ $(i \neq j)$.
Then a variable $x_j^{(g)}$ is exogenous for any $g$ if and only if the following holds for any $g$: $x_j^{(g)}$ is independent of its residuals $r^{(g)}_{i,(j)}$ for all $i \neq j$. \mbox{\hfill \qed}
\end{lem} 

\paragraph{Proof}
(i) Assume that $x_j^{(g)}$ is exogenous for any $g$. 
Applying Lemma~\ref{lemma1} on the LiNGAM of each group $g$, $x_j^{(g)}$ is independent of its residuals $r^{(g)}_{i,(j)}$ for all $i \neq j$.
(ii) Assume that $x_j^{(g)}$ is not exogenous for some $g$. 
For the $g$, there exists such $i$ that $x_j^{(g)}$ is not independent of its residual $r^{(g)}_{i,(j)}$ because of Lemma~\ref{lemma1}. 
From (i) and (ii), the lemma is proven.  \mbox{\hfill \bsquare} 

Thus, to estimate a variable that is exogenous in every group, we propose to find the variable subscript of such a variable that minimizes the weighted sum of the individual independence measures in Eq.~(\ref{eq:Tkernel}): 
\begin{eqnarray}
T(j; U) & = & \sum_{g=1}^{c} w^{(g)}T_{kernel}(x^{(g)}_j; U), \label{eq:T}
\end{eqnarray}
where $w^{(g)}$ is the non-negative weight of Group $g$. Here, we simply take the sample size of Group $g$ which we denote by $n^{(g)}$ as the weight $w^{(g)}$, that is, $w^{(g)}=n^{(g)}$. 
Then in each group we subtract the effect of the exogenous variable found from the other variables using least squares regression.  Similarly to the single group case, we can find all the causal orders by iterating this.

This is a deflation approach and enables estimation of sub-orderings instead of the whole ordering. 
It would be a more practical goal to estimate the first $q$ ($q < p$ and $q< \min(n^{(1)},\cdots, n^{(c)}$)) orders  in a causal ordering rather than all the orders when the sample size is not very large compared to the number of variables \cite{Sogawa10ICANN,Hyva10ACML} since the number of causal orders to be estimated gets smaller.  Moreover, it is necessary to stop the algorithm before the covariance matrices of residuals get singular when any of the sample sizes is smaller than the number of variables. 

We now present a direct algorithm to estimate a causal ordering and the connection strengths in the multiple LiNGAMs (\ref{eq:model}):

\noindent
  %\pagebreak
  \rule{\columnwidth}{0.5mm}
       { \sffamily
\vspace{-4mm}
	 \begin{enumerate}
	 \item Given a set of $p$-dimensional random vectors $\bfx^{(g)}$ ($g=1,\cdots,c$), a set of its variable subscripts $U$ and $p \times n^{(g)}$ data matrices of the random vectors as $\X^{(g)}$ respectively, initialize an ordered list of variables $K:=\emptyset$ and $m:=1$ and choose the number of causal orders to be estimated $q$ ($q=1, \cdots, {\rm or}\ p$). 
	 \item Remove the mean from the data. 
	 \item Repeat until $\min(q,p-1)$  subscripts are appended to $K$:
	 \begin{enumerate}
	 \item \label{step:2a} For each group $g$, perform least squares regressions of $x_i^{(g)}$ on $x_j^{(g)}$ for all $i \in U \backslash  K$ ($i \neq j$) and compute the residual vectors $\bfr^{(g)}_{(j)}$ and the residual data matrix $\R^{(g)}_{(j)}$ from the data matrix $\X^{(g)}$ for all $j \in U \backslash  K$. Find the variable subscript $m$ of such a variable that is most independent of its residuals:  	 
	\begin{eqnarray}
	m=\arg\min_{j \in U \backslash  K} T( j; U \backslash  K), 
	\end{eqnarray}
	where $T$ is the independence measure defined in Eq.~(\ref{eq:T}). 
	\item Append $m$ to the end of $K$. 
	\item Let $\X^{(g)}:=\R^{(g)}_{(m)}$ and replace $\bfx^{(g)}$ by $\bfr^{(g)}_{(m)}$, that is, $\bfx^{(g)}:=\bfr^{(g)}_{(m)}$ ($g=1,\cdots,c$). 
	 \end{enumerate}
	 \item Append the remaining variable to the end of $K$ if $q=p$.
	 \item \label{step:reg} For each group g, construct a $q \times q$ strictly lower triangular matrix $\B^{(g)}_q$, that is, a sub-matrix of $\B^{(g)}$ by following the order in $K$, and estimate the connection strengths by using least squares regression on the original random vectors $\bfx^{(g)}$ and the original data matrix $\X^{(g)}$. 
	 \end{enumerate}
       } \vspace{-4mm}
\noindent \rule{\columnwidth}{0.5mm}
This essentially reduces to the original DirectLiNGAM \cite{Shimizu11JMLR} if the number of groups is one, that is,  $c=1$. 

%%%%%%%%%%%%%%%%%%%%%%%%%%%%%%%%%%%%
\section{Simulations}\label{sec:simulations}
%%%%%%%%%%%%%%%%%%%%%%%%%%%%%%%%%%%%
As a sanity check of our method, we performed two experiments with simulated data. 
The first experiment consisted of 100 trials. 
In each trial, we generated $10$ datasets with dimension $p=10$ and sample sizes  $n^{(1)}=\cdots=n^{(5)}=50$ and $n^{(6)}=\cdots=n^{(10)}=100$ and applied our joint estimation method on the data. 
For comparison,  we also tested two methods; 1)  a separate estimation method that applies DirectLiNGAM \cite{Shimizu11JMLR} on the datasets separately and 2) a naive method that creates a single dataset concatenating all the datasets and applies DirectLiNGAM on it. 
%:
The sample sizes would not be very sufficient to reliably estimate the models separately \cite{Sogawa10IJCNN}. 
Each dataset was created as follows: 
\begin{enumerate}
\item We constructed the $p \times p$ connection strength matrix with all zeros and replaced every element in the lower-triangular part by independent realizations of Bernoulli random variables with success probability $s$ similarly to \cite{Kalisch07JMLR}. The probability $s$ determines the sparseness of the model. We set the sparseness $s$ so that the expected number of adjacent variables was half of the dimension. 
\item We replaced each non-zero entry in the connection strength matrix by a value randomly chosen from the interval $[-1.5, -0.5]$ $\cup$ $[0.5, 1.5]$ and selected variances of the external influences from the interval $[1, 3]$. 
The resulting matrix was used as the data-generating connection strength matrix $\B^{(g)}$. 
\item \label{step:2} We generated data with sample size $n^{(g)}$ by independently drawing the external influence variables $e_i^{(g)}$ from various 18 non-Gaussian distributions used in \cite{Bach02JMLR} including super- and sub-Gaussian distributions and symmetric and asymmetric distributions \cite{Shimizu11JMLR}. 
\item The values of the observed variables $x_i^{(g)}$ were generated using the connection strength matrix $\B^{(g)}$ and external influences $e_i^{(g)}$. Then, we generated constants following the Gaussian distribution $N(0,4)$ and added them to the variables as their means. 
Finally, we permuted the variables according to a random ordering shared by all the groups. 
\end{enumerate}
 
We computed the percentages of datasets where {\it all} the causal orders were correctly learned in the total datasets, that is, 1,000 datasets ($=$ 10 datasets $\times$ 100 trials).  
The percentages for the joint estimation method, the separate estimation method and the naive method were 96.6\%,  44.9\% and 0.4\%, respectively. 
Further, we computed the average squared errors between true connection strengths and estimated ones. 
The average squared errors for the three methods were 0.02,  0.07 and 0.51, respectively. 

In the second experiment, we considered a situation with more variables than observations, that is,  $p>n^{(g)}$. 
We generated $10$ datasets with dimension $p=40$ and sample size $n^{(1)}=\cdots=n^{(5)}=10$ and $n^{(6)}=\cdots=n^{(10)}=20$ in the same manner as above and only estimated the first five variables in a causal ordering ($q=5$). 
We repeated this procedure 100 times. 
The percentages of datasets for which the joint, separate and naive methods correctly estimated such first five variables were 92.1\%, 43.1\% and 44.4\%, respectively. 
The average squared errors for the three methods were 0.09, 0.21 and 0.33, respectively. 

\begin{figure}[!tb]
  \begin{center}
  \begin{minipage}{.45\linewidth}
\includegraphics[width=3.00in]{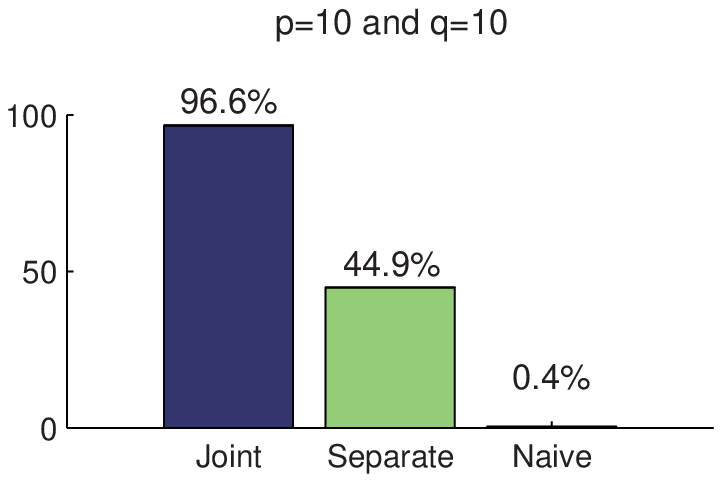}
  \end{minipage}
  \hspace{1.0pc}
  \begin{minipage}{.45\linewidth}
\includegraphics[width=3.00in]{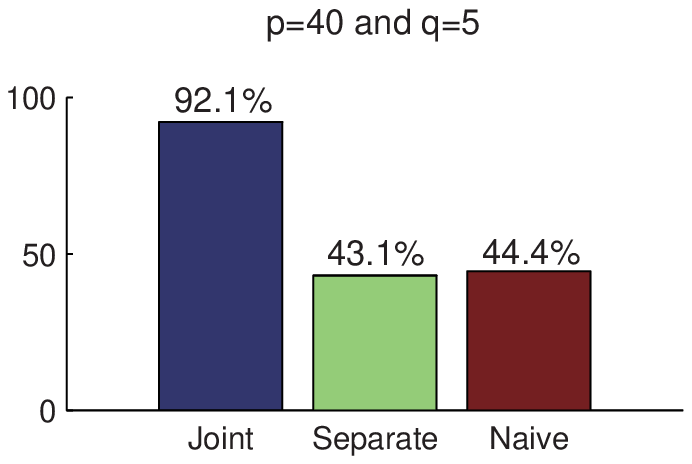}
  \end{minipage}
  \end{center}
\caption{Percentages of datasets with successful causal order discovery for the three methods: our joint method, the separate method \cite{Shimizu11JMLR} and the naive method. Left: $p=10$ and $n^{(1)}=\cdots=n^{(5)}=50$ and $n^{(6)}=\cdots=n^{(10)}=100$. All the causal orders were estimated ($q=10$). Right: $p=40$ and $n^{(1)}=\cdots=n^{(5)}=10$ and $n^{(6)}=\cdots=n^{(10)}=20$. First five causal orders were estimated ($q=5$).}
  \label{fig:results}
  \end{figure}

To sum up, our joint estimation method worked much better than the other two methods in the simulations as summarized in Fig.~\ref{fig:results}.   

%%%%%%%%%%%%%%%%%%%%%%%%%%%%%%%%%%%%
\section{Conclusions}\label{sec:conc}
%%%%%%%%%%%%%%%%%%%%%%%%%%%%%%%%%%%%
We proposed a new joint estimation method for multiple LiNGAMs. 
The new method assumes that such LiNGAMs share a causal ordering, but the connection strengths of variables and the distributions of external influences may be different between the models. 

%%%%%%%%%%%%%%%%%%%%%%%%%%%%%%%%%%%%
\section*{Acknowledgements}
We would like to thank Yoshinobu Kawahara, Takashi Washio, Satoshi Hara and Aapo Hyv\"arinen for interesting discussions.
This work was supported by MEXT Grant-in-Aid for Young Scientists \#21700302.

%%%%%%%%%%%%%%%%%%%%%%%%%%%%%%%%%%%%%
%\appendix

% References
\bibliography{shimizu11b_arXiv}
\bibliographystyle{unsrt}

\end{document}